\pdfoutput=1

\documentclass[11pt]{article}

\usepackage[]{coling}

\usepackage{times}
\usepackage{latexsym}

\usepackage[T1]{fontenc}

\usepackage[utf8]{inputenc}

\usepackage{microtype}

\usepackage{inconsolata}

\usepackage{graphicx}

\usepackage{amssymb}
\usepackage{xurl}
\usepackage{subcaption}

\title{Comparable Corpora: Opportunities for New Research Directions}

\author{Kenneth Church \\
Northeastern University \\
  \texttt{k.church@northeastern.edu} \\}

\begin{document}
\maketitle
\begin{abstract}
Most conference papers present new results, but this paper will focus more on opportunities for the audience to make their own contributions.  This paper
is intended to challenge the community to think more broadly about what we can do with comparable corpora.
We will start with a review of the history, and then suggest new directions for future research.
\end{abstract}

\section{Introduction}

The success of chat bots in many languages demonstrates the power of comparable corpora (CC) and pivoting via English.
We will start with a review of the history of CC, and then suggest new directions for future research:

\begin{enumerate}
    \setlength{\itemsep}{0pt}
    \setlength{\parskip}{0pt}
    \setlength{\parsep}{0pt}
        \item More depth: CC are normally used for simple tasks such as bilingual lexicon induction (BLI),
    but CC can be used for much more interesting views of lexical semantics.
    \item Compare and Contrast: CC are normally used to make simple comparisons over language pairs, but they can be used for contrasts as well as comparisons (in monolingual settings as well as multilingual settings).
    \item More modalities: Now that vectors encode everything (text in many languages, pictures, audio, video), we can compare and contrast everything with everything.
      \item Bursting filter bubbles:
    bots made in America are trained on corpora from an American perspective with American biases.
    We should not impose American values on others.
\end{enumerate}



\section{Historical Background}
\label{sec:history}

\begin{table}[hb!]
   \centering
  \begin{tabular}{l r }
  \textbf{Resource} & \textbf{Cites} \\ \hline
  Europarl \cite{koehn-2005-europarl} & 4634 \\
  OPUS \cite{tiedemann-2012-parallel} & 2255 \\
  ~~~HF: Helsinki-NLP/opus-100 \\
  \cite{resnik-smith-2003-web} & 848 \\
 MultiUN \cite{eisele2010multiun} & 327 \\
 ~~~HF: {\small{Helsinki-NLP/multiun}} \\
 Bible \cite{pratap2024scaling} & 254 \\
  ~~~\cite{,akerman2023ebible} \\
  ~~~ HF: {\small{Flux9665/BibleMMS}}  \\
  \href{https://catalog.ldc.upenn.edu/LDC95T20}{LDC: Hansard French/English} & \\
  \href{https://catalog.ldc.upenn.edu/LDC2000T50}{LDC: Hong Kong Hansards} & \\
  HF: {\small{NilanE/ParallelFiction-Ja\_En-100k}} \\
  HF: {\small{sentence-transformers/parallel-sentences}} \\
  HF: {\small{tiagoblima/bible-ptbr-gun-gub-aligned}} \\
  HF: {\small{dsfsi/vukuzenzele-sentence-aligned}} \\
  ~~~\cite{marivate_vukosi_2023_7598540} \\
    \end{tabular}
      \caption{Examples of parallel corpora}
  \label{tab:parallel}
\end{table}


\begin{table}
   \centering
  \begin{tabular}{l l l }
\textbf{English} & \textbf{French} & \textbf{Sense} \\ \hline
bank & banque & money \\
  & banc & river \\ \hline
  
duty  & droit &  tax \\
  & devoir & obligation \\ \hline

drug & m\'edicament & medical \\
& drogue & illicit \\ \hline

land & terre & property \\
 & pays & country \\ \hline

language & langue & medium \\
& langage & style \\ \hline

position & position & place \\
 & poste & job \\ \hline

sentence & peine & judicial \\
 & phrase & grammatical \\ \hline
 
    \end{tabular}
      \caption{Using Hansards for Word Sense Disambiguation (WSD), based on Table 2 in \citet{gale1992method}}
  \label{tab:WSD_Hansards}
\end{table}

\subsection{Parallel Corpora}

\autoref{tab:parallel} shows some examples of parallel corpora.
HF and LDC in \autoref{tab:parallel}
refer to HuggingFace\footnote{\url{https://huggingface.co/}} and the Linguistic Data Consortium,\footnote{\url{https://www.ldc.upenn.edu/}} respectively.
More parallel corpora can be found on HF by searching for \textit{parallel}, \textit{aligned} and \textit{translation}.

The main application of parallel corpora has been machine translation \cite{brown-etal-1993-mathematics}.
Shannon's noisy channel model \cite{shannon1948mathematical} was originally motivated for
applications in communication (telephones), but
it has been used for many other applications including
machine translation.
That is, to translate from English to French,
one imagines that French speakers think in English,
like English speakers do, but for some reason, when French speakers talk,
the noisy channel converts their English to French.  

\begin{equation}
\label{eqn:EF}
    E \rightarrow Noisy~Channel \rightarrow F
\end{equation}
\noindent

\noindent
The task of
the translation system is to recover the original English, $E$, from the observed French,
$F$.  
These days, it has become standard practice to use neural networks
for translation, but it used to be popular to use Hidden Markov Models (HMMs)
to find the most likely English, $\hat{E}$,
based on
a prior (language model), $Pr(E)$,
and a bilingual dictionary, $Pr(F|E)$.

\begin{equation}
\label{eqn:EF2}
  \hat{E} = \mathrm{argmax}_E Pr(E) Pr(F|E)
\end{equation}

\noindent
Much of the discussion below will focus on the
bilingual lexicon, $Pr(F|E)$.  $Pr(E)$, the language model in Eqn~(\ref{eqn:EF2}),
is relatively well estimated because we can re-use monolingual
LLMs that have been developed for other applications.
The bilingual
lexicon, $Pr(F|E)$, assigns probabilities
to all sequences of English, $E$, and French, $F$.
It was standard practice, at least at first,
to estimate $Pr(F|E)$ from parallel corpora such
as the English-French Canadian Hansards.

The rest of this section on history will largely focus on the lexicon.
After introducing comparable corpora as an alternative to parallel corpora,
we will motivate WSD (word-sense disambiguation).  Much of the research on WSD started with bilingual word-senses, but it should be noted
that word-senses are different in 
monolingual and bilingual dictionaries.

There has also been considerable work on transferring monolingual
lexical resources such as WordNet and VAD to more languages.
Unfortunately, much of this work uses translation to pivot out of English in inappropriate ways,
as we will see.  

This section will end with a review of BLI (bilingual lexicon induction).
The BLI literature uses more modern methods in machine learning than previous methods for inducing lexicons from CC,
but BLI benchmarks (such as MUSE) may not be as effective as older WSD methods
for addressing classic challenges with translations of ambiguous words.
A classic example is \textit{bank},
which is translated as \textit{banque} and \textit{banc} in the Canadian Hansards, depending
on the sense.
Unfortunately, this ambiguity is not captured in the MUSE benchmark
where \textit{bank} translates to \textit{banque} (but not \textit{banc}).  In the reverse direction, MUSE has translations for both \textit{banque} and \textit{banc},
but they translate to different English words, \textit{bank} and  \textit{bench}, respectively.  Comparisons of ambiguities in Hansards (\autoref{tab:WSD_Hansards}) and MUSE (\autoref{tab:MUSE} and \autoref{tab:MUSE2}) suggest that MUSE is not testing WSD as much as the older literature.  Another concern with MUSE is that most words in the benchmark translate to themselves.  These concerns suggest that there
may be room to introduce a new benchmark that would make a stronger case
for comparable corpora (CC).

After discussing history, the next section will discuss more radical challenges for
the future: lexical semantics, transfer learning, filter bubbles and connections between academic search and CC.

\subsection{Comparable Corpora (CC)}


The term, \textit{comparable corpora}, was introduced  in \cite{fung-church-1994-k,rapp-1995-identifying,fung-yee-1998-ir-approach,fung2000statistical} to address limitations with
parallel corpora.  
Parallel corpora are available for a few genres such as parliamentary debates (Hansards)
and religion (Bible), as shown in \autoref{tab:parallel}.
Since most texts and most genres
are not translated, we can collect larger and more diverse
corpora if we relax the restriction on translation.  CC replace
a single parallel corpus with two monolingual corpora, ideally
on similar (comparable) topics.


\subsection{Word-Sense Disambiguation (WSD)}

In addition to machine translation applications mentioned above, parallel corpora have also been used to disambiguate ambiguous words such as \textit{bank}, as illustrated in \autoref{tab:WSD_Hansards}.  \citet{bar1960present} thought machine translation
was impossible when
he could not figure out how to disambiguate words such as those in \autoref{tab:WSD_Hansards}.  It was
obvious that the translation depends on a solution to WSD.  

\citet{gale1992method} used this argument in reverse to obtain large quantities of labeled text
for WSD research.  They used parallel corpora such as Hansards to find instances
of ambiguous words such as \textit{bank}, and use the French translations to label each instance of \textit{bank} as either ``money'' sense or ``river'' sense.  After labeling the English in this way, they threw away the French and used the sense-labeled text to train and test machine learning
methods for WSD.

\subsection{Monolingual Senses != Bilingual Senses}
\label{sec:mono_neq_bilingual}

This approach was successful in reviving interest in WSD research, though
it should be mentioned that bilingual lexicography is different from
monolingual lexicography.  Consider the word \textit{interest}.  This  word has many senses including a ``money'' sense and a ``love'' sense, among others.
A monolingual dictionary will describe each of these senses in considerable detail.
However, there will be little to say about \textit{interest} in an English-French bilingual dictionary because
the same complications are shared between the English word and its French equivalent.
Thus, the approach above is more effective for
words like those in \autoref{tab:WSD_Hansards} where the word is ambiguous in one language but not the other, and less effective for words like \textit{interest}, which are equally ambiguous in both
languages.

\subsection{Inappropriate Uses of Translation}

Parallel corpora are limited in a number of ways.  Genre is perhaps the most
obvious limitation, but a more serious limitation may be distortions
introduced by translation.  

When I was first working with Hansards in the 1990s, I tried to pitch parallel corpora to Sue Atkins, a lexicographer who specialized in English-French bilingual dictionaries.  She rejected my pitch, objecting to ``translationese''\footnote{\url{https://en.wiktionary.org/wiki/translatese}}
as ``unnatural'' natural language.  In addition, she criticized concordance tools for parallel corpora
because they failed to distinguish source and target languages.
Examples of these tools can be found in the sketch engine;\footnote{\url{https://www.sketchengine.eu/guide/parallel-concordance-searching-translations/}}
these tools show examples of a word in one language as well as its equivalents in other languages.


\subsubsection{XNLI: A Multilingual version of NLI}

Much of the work on parallel corpora treats the source and target languages
as equivalent (with equal status), ignoring distortions introduced by translation.
We should be more careful about translation artifacts in many benchmarks.  \citet{artetxe-etal-2020-translation}
call out XNLI, a English version of an NLI task.
The monolingual NLI task depends on word overlaps between the premise and the hypothesis,
but many of these crucial overlaps are lost in translation in the XNLI version where
premises and hypotheses are translated independently.
Too much of the work in computational linguistics uses translation to pivot via English in inappropriate ways.

\subsubsection{No Language Left Behind (NLLB)}

\citet{abdulmumin-etal-2024-correcting} report serious problems with FLORES \cite{goyal-etal-2022-flores}
in four African languages.
A common problem was the use of Google Translate, which sometimes produced ``incoherent or
unclear'' Hausa text.
FLORES is an important test set for NLLB (no language left behind) \cite{team2022NoLL}.

\begin{table}
   \centering
  \begin{tabular}{l l }
\textbf{Synset} & \textbf{French Glosses} \\ \hline
dog.n.01 & canis\_familiaris, chien \\
cat.n.01 & chat\\
house.n.01 & maison \\
bank.n.01 & banque, rive \\ \hline
    \end{tabular}
      \caption{Global WordNet pivots from English}
  \label{tab:wordnet}
\end{table}

\begin{table}
   \centering
  \begin{tabular}{l l l l l}
\textbf{English} & \textbf{Hausa} & \textbf{V} & \textbf{A} & \textbf{D} \\ \hline
aaaaaaah & aaaaaaa & 0.48 & 0.61 & 0.29 \\
aaaah & aaaah & 0.52 & 0.64 & 0.28 \\
aardvark & ardvark & 0.43 & 0.49 & 0.44 \\
aback & abin mamaki & 0.39 & 0.41 & 0.29 \\
abacus & abacus & 0.51 & 0.28 & 0.49 \\
abalone & abalone & 0.50 & 0.48 & 0.41 \\ \hline
    \end{tabular}
      \caption{NRC-VAD pivots from English using Google Translate;
      V = Valance, A = Arousal \& D = Dominance}
  \label{tab:vad}
\end{table}

\subsubsection{WordNet and VAD}

\autoref{tab:wordnet} and \autoref{tab:vad}
show attempts to use translation to pivot
from English to other languages.  WordNet\footnote{\url{https://www.nltk.org/howto/wordnet.html}} \cite{miller1995wordnet}
and NRC-VAD \cite{mohammad-2018-obtaining}\footnote{\url{https://saifmohammad.com/WebPages/nrc-vad.html}} were originally designed for English.
Translation was used to transfer them to more languages.
Note that translation introduces losses; bank.n.01 cannot be both the money sense (\textit{banque}) and the river sense (\textit{rive}).
I asked a colleague, a native speaker of Hausa,
to comment on \autoref{tab:vad}.  None of the Hausa words
in the table are that useful.  Most of the words
in the Hausa column are English, with the exception of \textit{abin mamaki} which Google
translates to \textit{what a surprise} in English.  My informant did not know
what \textit{aback} means in English even though his English is excellent.  When I explained
it to him, we agreed that this translation is not convincing.

\begin{table}[ht!]
   \centering
  \begin{tabular}{l l  }
\textbf{English} & \textbf{French}  \\ \hline
bank & banque, banques, \textit{but not} banc \\
duty & devoir, \textit{but not} droit\\
drug & drogue, médicament\\
land & terre, terrain, terres, \textit{but not} pays \\
language & langue, langues, langage \\
position &  position, \textit{but not} post\\
sentence & peine, phrase, sentence \\
good & bien, bon, bonne, bonnes, bon \\
bad & mal, mauvais, mauvaise, bad \\
 \hline
     \end{tabular}
      \caption{Some examples from MUSE: fr $\rightarrow$ en}
  \label{tab:MUSE}
\end{table}

In short, there are many problems with using translation to pivot from
English to many other languages.  It is unlikely that the structure of the English WordNet ontology
and the English VAD lexicon is universal over all languages.  
In the West, we slay dragons, but in the East, dragons are good luck.
In the West, white is common for weddings and black is common for funerals, but
in some places, white is common for funerals, and in other places, red is common for weddings.
Even the list of concepts is likely
to vary from one language to another.  Many of the English words in \autoref{tab:vad} are not (much of) ``a thing'' 
in Hausa.

\begin{table}[ht!]
   \centering
  \begin{tabular}{l l  }
\textbf{French} & \textbf{English}  \\ \hline
banc & bench, \textit{but not} bank \\
banque & bank, banking \\
droit & right, law \\

devoir & duty \\

drogue & drug, drugs, drogue\\

médicament & medicine, drug, medication\\

terre & land, earth, soil, terre\\

terrain & land, terrain\\

terres & land, lands\\

langue & language\\

langues & language, languages \\

langage & language\\

position & position\\

peine & sentence, pain, penalty, sorrow\\

phrase & sentence, phrase \\

sentence & sentence, sentencing\\
 \hline
     \end{tabular}
      \caption{Some examples from MUSE: en $\rightarrow$ fr}
  \label{tab:MUSE2}
\end{table}

\begin{table}[ht!]
   \centering
   \setlength{\tabcolsep}{5pt}
  \begin{tabular}{c c c c c }
\textbf{Dict} & \textbf{Pairs} & \textbf{Src} & \textbf{Tgt} & \textbf{Src=Tgt} \\ \hline
${\mathrm{en} \rightarrow \mathrm{fr}}$ & 113,286 & 94,681 & 97,035 & 73,471 \\
${\mathrm{fr} \rightarrow \mathrm{en}}$&  113,324 & 97,021 &  94,730 & 73,471 \\
 \hline
     \end{tabular}
      \caption{MUSE Dictionary Sizes}
  \label{tab:MUSE_sizes}
\end{table}

\subsection{Bilingual Lexicon Induction (BLI)}

Much of the work on BLI is based on the MUSE benchmark \footnote{\url{https://github.com/facebookresearch/MUSE}}
\cite{lample2017unsupervised,conneau2017word}.
The MUSE benchmark provides:
\begin{enumerate}
    \setlength{\itemsep}{0pt}
    \setlength{\parskip}{0pt}
    \setlength{\parsep}{0pt}
    \item fastText\footnote{\url{https://github.com/facebookresearch/fastText}} embeddings for 30 languages, and
    \item gold set of bilingual dictionaries, $D_{l_i \rightarrow l_j}$ for 110 pairs of languages: $l_i, l_j$.  The gold sets are split into training (seed) dictionaries and test dictionaries.
\end{enumerate}

\noindent
See section 2.2 of \cite{sharoff2023bucc} for an introduction
to vector space models and CC.
The fastText embeddings, $X_l \in \mathbb{R}^{|V_l| \times d}$,
contain a row for each word in the vocabulary, $V_l$, for language $l$.
The rows are vectors of length $d$, where $d$ is the number of hidden
dimensions.

Each dictionary, $D_{l_i \rightarrow l_j}$ consist of a list of pairs of words in the two languages.
\autoref{tab:MUSE_sizes} counts the number of pairs in both directions,
as well as the number of unique words in the source language (src) and target
language (tgt).  Many of the pairs use the same word in both languages, as indicated by the last column.


The task is to estimate a dictionary, $\hat{D}_{l_i \rightarrow l_j}$, for a pair of languages, $l_i$ and $l_j$.  We then compare estimates, $\hat{D}$, with gold dictionaries, $D$.
A simple approach is to use the training (seed) dictionaries to estimate a rotation matrix, $R \in \mathbb{R}^{d \times d}$, where $R = \mathrm{argmin}_R ~ || R X_{l_i} - X_{l_j}||_F^2$.  It is standard practice to estimate $R$ with the
orthogonal Procrustes problem\footnote{\url{https://docs.scipy.org/doc/scipy/reference/generated/scipy.linalg.orthogonal_procrustes.html}} \cite{schonemann1966generalized}.

At inference time, we start with a vector in $X_{l_i}$,
and then rotate those vectors by $R$ and use approximate nearest neighbors (ANN) \cite{bruch2024foundations}
to find nearby vectors in $X_{l_j}$.

Early work on CCs attempted to collect word lists \cite{kilgarriff2014lrej} and infer bilingual lexicons; MUSE
updates this approach using modern methods in machine
learning.  That said, MUSE may not be as effective
as older methods for WSD because of gaps.
Examples from $D_{\mathrm{fr} \rightarrow \mathrm{en}}$
and $D_{\mathrm{en} \rightarrow \mathrm{fr}}$ are shown in Tables \ref{tab:MUSE}-\ref{tab:MUSE2}; some of the ambiguities
in \autoref{tab:WSD_Hansards} are covered, and some are not.  An example
of a gap is: \textit{bank} (en) $\rightarrow$ \textit{banc} (fr);
this pair is missing from both $D_{\mathrm{en} \rightarrow \mathrm{fr}}$  and $D_{\mathrm{fr} \rightarrow \mathrm{en}}$.


\section{Challenges for the Future}
\label{sec:future}

\subsection{BLI, PMI and Lexical Semantics}

Much of the
work on BLI uses a simple view of a bilingual lexicon where single words in one language
correspond to single words in another language, more or less one-for-one.
Obviously, the relationship is far more complicated than this.  The phrasal verb, \textit{ask for}, is similar
to \textit{request}, violating the one word for one word assumption.

\subsubsection{Etymology}

More seriously, there is a difference in register, going back to the Norman Conquest in 1066. 
For a few hundred years after 1066, the English Court spoke French.  As a result, English borrowed many words from French.  The French term
typically has a higher register than the older English equivalent; 
the peasants raise \textit{cows, calf} and \textit{swine} so the aristocracy can eat \textit{beef, veal} and \textit{pork}.\footnote{\url{https://www.csmonitor.com/The-Culture/In-a-Word/2021/0510/They-re-cows-in-the-field-but-beef-on-the-table}}  

\subsubsection{Distributional Methods}

Much of the work on BLI does not take advantage of etymology because  work on BLI is based on the Distributional Hypothesis\footnote{\url{https://aclweb.org/aclwiki/Distributional_Hypothesis}} \cite{harris1964distributional}
and Firth's ``You shall know a word by the company it keeps'' \cite{firth1957synopsis}.
The distributional hypothesis is convenient for computation, suggesting ``(unlabeled) corpora are all we need,'' though many aspects of linguistics go beyond distributional evidence, e.g., etymology, lexical semantics.

There are interesting connections between popular distributional methods, e.g.,
PMI (pointwise mutual information), Word2Vec and LLMs (large language models).  The connection between BLI and Word2Vec was
mentioned above.
\citet{levy2014neural} view Word2Vec as a factored representation of PMI \cite{church-hanks-1990-word}. 
BERT and chat bots can be viewed as an enhancement of Word2Vec; instead
of representing words as vectors, we now represent sequences of 512-subword units as vectors.

\subsubsection{Lexical Semantics}

As mentioned above, lexical semantics is a challenge for distributional
methods.
While there are some similarities between PMI (collocations) and
lexical semantics (synonyms, antonyms, is-a), there are also some important differences, as shown in \autoref{tab:semantics}.
PMI scores are large when words appear near one another more than chance.
Consequently, both synonyms and antonyms have large PMI scores because
documents often compare and contrast this with that.
Similarly, PMI scores can be large for other words that appear near
one another, e.g., 
\textit{window, door} and \textit{house}.  Large PMI scores do not necessarily imply synonymy.

\begin{table}
   \centering
  \setlength{\tabcolsep}{2pt}
  \begin{tabular}{ r | l l l }
\textbf{Relation}     & \textbf{PMI} & \textbf{Lexical Sem} & \textbf{Back Trans} \\ \hline
Synonyms & large & $=$ (equiv. rel.) & large\\
Antonyms & large & $\neq$ (anti-sym) & small \\
Is-a     & small & $\leq$ (partial order) & small \\
Part-Whole & large &  & small \\ \hline 
  \end{tabular}
      \caption{PMI $\neq$ Lexical Semantics}
  \label{tab:semantics}
\end{table}

Back translations are also mentioned in \autoref{tab:semantics}.
Back translations are more effective than PMI for distinguishing synonyms from antonyms.
If we take a random walk over MUSE dictionaries and start from \textit{good}, such walks will often take us to synonyms, but rarely to antonyms.
There is an opportunity to propose a theory of translation and collocation
based on linear algebra and graph theory.  This theory should explain the observations
in \autoref{tab:semantics} where antonyms are close in terms of PMI but not in
terms of random walks on translations.

\begin{table*}
   \centering
  {\small
  \begin{tabular}{c r c r r r r r}
\textbf{Language} & \textbf{\href{https://en.wikipedia.org/wiki/List_of_Wikipedias}{Wikipedia}} & \textbf{Joshi} & \textbf{S2 Abstracts} & \textbf{ACL} &\textbf{ HF Datasets} &\textbf{ HF Models} & \textbf{Speakers}\\ \hline

en	& 6,917,939 & 	5	& 88,348,938 & 	103,000 & 	10,749	& 50,717 & 1456M \\

zh & 1,452,669 & 	5	&  3,061,847 & 71,800 &	1202 &	4495	& 1138M \\

hi	& 163,524 &	4	&	2,848  & 8,740 &	421	& 1388	& 610M \\

es	& 1,992,685 &	5	& 2,742,468  &	28,600	& 945	& 3245 &	559M\\

fr	& 2,650,236	& 5	&2,772,266	& 35,500 & 1064 & 	4033	& 310M\\

id	& 711,624	& 3 & 	2,234,953	& 4,230 & 	395 &	1317	& 290M\\ 

ar	& 1,625,651	& 5	& 149,043	& 17,900&	558&	1681&	274M\\

bn & 	160,408	& 3	& 445& 3,270 & 	298 &	788	& 273M\\

pt	& 1,138,923	& 4	& 1,937.959 &	9,660 & 	596	& 1935 &	264M\\

ru	& 2,012,648	& 4	& 509,503  &	13,300 &	799	& 2307 &	255M\\

ur &	215,081	& 3 & 	454	& 3,220 & 	204 &	658	& 232M\\

de &	2,964,125 &	5	& 1,227,473&	42,900	& 789	& 348	& 133M\\
ja	& 1,438,806 &	5 &	317,394  & 	38,200 &	596 &	2887	& 123M\\
mr	& 98,559 &	2	& 275	& 1,480 & 	193	& 642&	99M\\
te& 	101,681 &	1	& 13	& 2,120&	223 &	589&	96M\\
tr	& 624,742	& 4	& 370,727  &	8,490 &	398	& 1389 &	90M\\
ta & 	169,766 & 	3	& 728	& 3,980 & 	263 &	1030	& 87M\\ 
vi	&1,294,281&	4 &	44,477 &	3,010	& 474	& 1188	& 86M\\
tl &	47,891 &	3 &	933 &	1,100	& 116	&451	&83M\\
ko&	691,121 &	4&	793,921  &	16,900 &	534 &	2741	& 82M\\
ha &	51,659	& 2	& 	& 	823 & 98	& 441	& 79M\\ 
jv	& 74,159	& 1	& 	& 535 &	76	& 342	& 68M\\
it &	1,893,522  &		4 & 184,535	& 14,400 &	516	& 2129 &	68M\\
gu	& 30,474 &	1 &	23	& 263 &	174&	581&	62M\\
th	& 169,192 &	3	& 41,628& 	12,700 &	326	& 900 &	61M\\
kn	& 33,026 &	1	& 	143 &1540 &	178 &	534&	59M\\
am	& 15,374 &	2	&	96 & 1110 & 	117	& 493&	58M\\
yo&	34,080 &	2 & 	18	& 799  & 	123 & 	458 &	46M\\ \hline
  \end{tabular}}
      \caption{Some resources for transfer learning from high resource languages to growth opportunities}
  \label{tab:resources}
\end{table*}

\subsubsection{Avoid Pivoting via English}
As mentioned in \autoref{sec:mono_neq_bilingual}, monolingual lexicography is different
from bilingual lexicography.  For example, \textit{interest} has many senses
in monolingual dictionaries, but not in bilingual dictionaries.  \textit{Bank}
is ambiguous in English, but not in French.  Bilingual dictionaries become
interesting when the senses are not isomorphic.  \autoref{tab:wordnet} and \autoref{tab:vad} take an overly simplistic view of the structure of the lexicon where the ontology (and VAD values) are assumed to be universal.
Translating from English is likely to introduce distortions.  Can we
do better than pivoting via English?

\subsection{Transfer Learning}
\label{sec:transfer}

Suppose we want to transfer from a high resource language such as English
to growth opportunities such as Indonesian (id) and Hausa (ha).  
We prefer the term, \textit{growth}, over terms such as low resources
to refer to languages with more speakers than resources, such as many of the
languages in \autoref{tab:resources}.
\autoref{tab:resources} is sorted by the number of speakers.\footnote{\url{https://en.wikipedia.org/wiki/List_of_languages_by_total_number_of_speakers}}
The columns are based on:
\begin{itemize}
    \setlength{\itemsep}{0pt}
    \setlength{\parskip}{0pt}
    \setlength{\parsep}{0pt}
    \item Articles in Wikipedia\footnote{\url{https://en.wikipedia.org/wiki/List_of_Wikipedias}}
    \item Joshi classification\footnote{\url{https://microsoft.github.io/linguisticdiversity/assets/lang2tax.txt}}
\cite{joshi-etal-2020-state}
 \item Abstracts in Semantic Scholar (S2) 
 \item Articles in ACL Anthology\footnote{Based on searches such as \url{https://aclanthology.org/search/?q=hausa}}
 \item Datasets and Models in HuggingFace (HF)
\end{itemize}

\noindent
The good news is that we have more resources these days for growth languages
than we had for English when we started EMNLP in 1990s.
In addition to the resources in \autoref{tab:resources},
there is support for most of these languages in multilingual LLMs, Google Translate,
and No Language Left Behind (NLLB) \cite{team2022NoLL}.

How can we transfer between languages with more resources and languages with fewer resources?
The crux of the problem is to construct a comparable corpus of English and the growth language.
Given that, there are a number of well-established methods to train language models.

Many efforts
start by pivoting from English.  
That is, they use English documents
as the source text, and then translate from English to the growth opportunity.
Filter bubbles are a problem for this approach.  This approach will not learn
aspects of the low resource language that go beyond what is in the high resource language.

We suggest using translation in the reverse direction, as well as similarities based
on recommender technologies in academic search engines.
That is, we will start with source texts in the growth language such as Wikipedia articles
and academic papers in Semantic Scholar (S2) \cite{Wade2022TheSS}.  We can then find ``nearby'' English
by several means:
\begin{enumerate}
    \setlength{\itemsep}{0pt}
    \setlength{\parskip}{0pt}
    \setlength{\parsep}{0pt}
    \item Translation from growth language to English
    \item Similar in a BERT-like vector space using Specter vectors \cite{cohan-etal-2020-specter} from S2
    \item Similar in terms of random walks on citations
\end{enumerate}
\noindent
By starting with documents in the growth language, we avoid the filter bubble criticism above.
In addition, professional translators specialize in one direction and not the other.  They
prefer to translate into their stronger language than vice versa.  We suggest
similar logic applies to transfer learning.  It is better for systems that are stronger in English to translate into English than vice versa.




\subsection{Filter Bubbles: A Monolingual Use Case}
\subsubsection{Filter Bubbles in News and Academia}

There are opportunities for CC to burst filter bubbles, both in monolingual
and multilingual applications.
With the rise of social media and cable news, we all live in filter bubbles.  
You may remember EMNLP was in Hong Kong just before COVID.  I was interested in the coverage of demonstrations in Hong Kong.  The story was very simple in New York and in Beijing.  The two perspectives disagreed in many respects, of course, but they agreed on simplicity.  When I went to Hong Kong for EMNLP, I learned that the story was anything but simple.
In short, we all have a tendency to oversimplify the truth, especially about events that are far way, \textit{of which we know little},\footnote{\url{https://www.iwp.edu/articles/2023/04/18/a-quarrel-in-a-faraway-country-between-people-of-whom-we-know-nothing/}}
like the famous cover of the New Yorker magazine with a view of the world
from 9th avenue.\footnote{\url{https://en.wikipedia.org/wiki/View_of_the_World_from_9th_Avenue}}

Ground News
has created a business by helping people see their blind spots.\footnote{\url{https://ground.news/blindspotter/methodology}}  They track coverage
in a range of different news outlets, and report who is saying what.  Is this story covered more
by outlets on the left or by outlets on the right?

This is an excellent place to start, but the news is fragmented in many more dimensions than just left/right in America.
The conflict in Sryia, for example, overlays three dimensions: (1) America/Russia, (2) Sunni/Shia and (3) Turkey/Kurds.  More dimensions are more challenging.

Academic conflicts have even more dimensions.  
Each school of thought has its position, and its friends and foes.
In \cite{church2011pendulum},
I suggested the pendulum has been swinging back and forth between empiricism and rationalism every 20 years.
Here is a slightly updated version of that argument:

\begin{itemize}
    \setlength{\itemsep}{0pt}
    \setlength{\parskip}{0pt}
    \setlength{\parsep}{0pt}
    \item Empiricism I (1950s):\\    
    Shannon, Skinner, Harris, Firth
    \item Rationalism I (1970s):\\
    Chomsky, Minsky
    \item Empiricism II (1990s):\\
    IBM, AT\&T Bell Labs, EMNLP, WWW
    \item Empiricism III (2010s):\\
Deep networks, LLMs, chat bots, RAG
\end{itemize}

\noindent
Why is the gap around 20 years? One suggestion
involves the cliche that grandparents and grandchildren
have a natural alliance.  Each academic generation rebels against the their teachers.
Chomsky and Minsky rebelled against methods that were popular in the 1950s,
and my generation returned the favor by reviving those methods.
When we started EMNLP (Empirical Methods in Natural Language Processing),
the E-word was an act of rebellion.

\begin{figure*}
          \centering
          \begin{subfigure}[b]{0.3\textwidth}       
  \includegraphics[width=1\columnwidth]{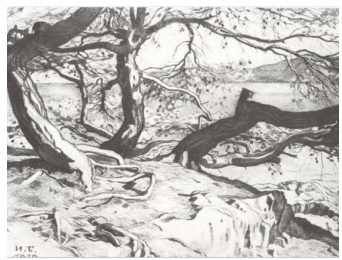}
  \caption{This tree is labeled ``sad'' in English and Chinese,
  but annotations in Arabic are more positive.}  
  \label{fig:tree}
           \end{subfigure}~\begin{subfigure}[b]{0.68\textwidth}       
    \includegraphics[width=1\columnwidth]{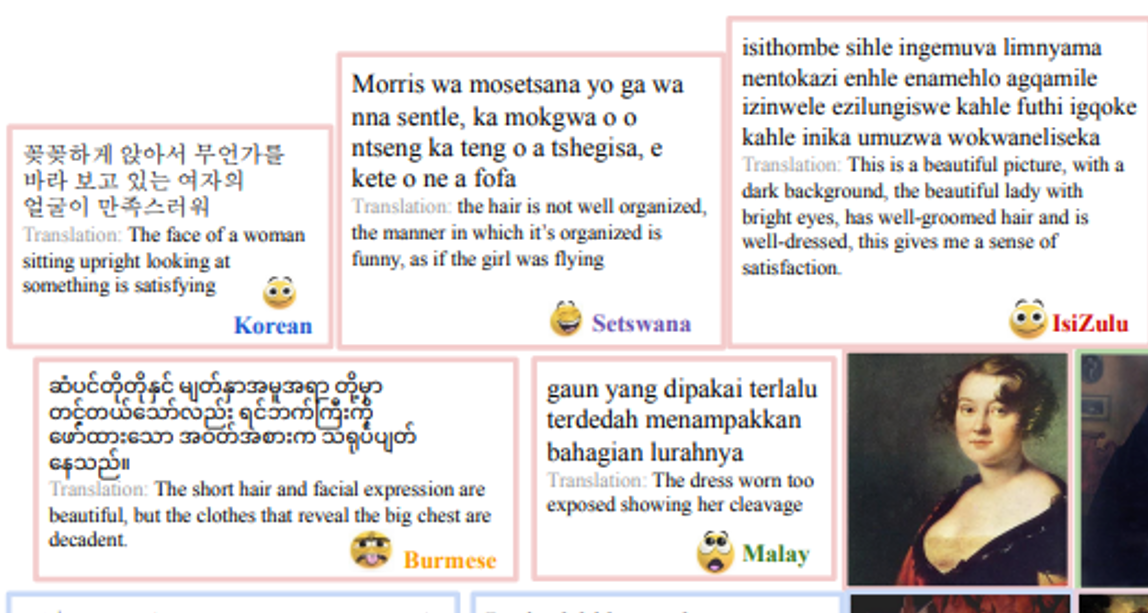}
  \caption{Many annotations are positive, but some object to the dress as too revealing.}  
  \label{fig:woman}
           \end{subfigure}
            \caption{\label{fig:artelingo} Emotion labels and captions depend on annotator's background (language/culture).} 
           \end{figure*}

\subsubsection{How can CC burst these filter bubbles?}

Suppose we consider Semantic Scholar to be a CC full of multiple
overlays that go well beyond Empiricism and Rationalism.
We can model the literature as schools of thought with agreements within clusters
and disagreements across clusters.



As suggested above, these days, it has become standard practice to
represent everything with vectors.  We can use vectors
to represent papers, as well as schools of thought.
Cosines can be used to estimate agreement and disagreement.
There are a number of ways to represent papers as vectors.  Two suggestions were mentioned above: BERT-like Specter vectors and random walks on citations\footnote{\url{https://github.com/VHRanger/nodevectors}} \cite{zhang2019prone}.
We normally use comparable corpora in bilingual applications,
but this application, clustering, has applications in both bilingual and monolingual settings.

\subsubsection{Comparable Corpora and Bots}

Web, news and social media offer many different perspectives
and points of view.  American bots
are trained on American corpora;
these bots currently lack a historian's ability to approach conflicts from multiple perspectives.

As homework for my NLP class \cite{church2024trust},
I asked students to write essays about the Opium War
from multiple perspectives including both
the East and the West.  They were encouraged to use
bots, but were told they would be responsible for
the content.  I had hoped students would rewrite
output from the bots, but few did.  
Even students from China handed in essays from an American
perspective, because 
American bots are trained on American corpora.  These
bots do not mention ``the century of humiliation,''\footnote{\url{https://www.uscc.gov/sites/default/files/3.10.11Kaufman.pdf}}
a perspective in the East which is motivating efforts to compete with
the West in AI so
China does not fall behind in technology like it did during
the Opium Wars.

Bot technology remains far behind historians like \citet{platt2019imperial}.
Bots see the world from a single (American) perspective.
Filter bubbles are dangerous; they contribute to trade wars and worse.

\subsection{Comparable Corpora and Pictures}

We normally think of corpora as text, but now that we are representing everything
as vectors, we can generalize corpora to include more modalities: text, speech, pictures, speech, video, etc.  As mentioned above, we are worried about pivoting from English prompts.  If we start
with English prompts, then we are likely to bias responses toward an English perspective.
\citet{mohamed-etal-2022-artelingo,mohamed-etal-2024-culture} starts with pictures
from WikiArt\footnote{\url{https://www.wikiart.org/}} as prompts.
Annotators are asked to add emotion labels and captions in 28 languages, as shown
in \autoref{fig:artelingo}.  Different annotators label pictures with different
emotion labels and captions, depending on their language and background.
The papers refer to a GitHub with a benchmark, as well as
baseline implementations of captioning systems that transfer from
high resource languages to growth opportunities.
Hopefully, the community will accept the challenge
and come up with even better systems that embrace diversity over many regions, cultures and languages.

It should be possible to beat a baseline system that translates the captions from English
to growth languages.  Consider the objections to the dress in \autoref{fig:woman}.
This is a case where it should be possible to outperform a captioning system that
translates from English because these objections are unlikely to be found in English
captions.
In fact, a reviewer asked for an ethics review, objecting
to the objections to the dress.  We are not siding with one
annotator over another, but we object to the objection to the objection.
It is not appropriate for us to impose American sensibilities on the rest
of the world.  Rather than remove biases from corpora (and WikiArt), we hope
to build bots that will be more aware of regional sensitivities to
topics such as: dress, nudes, religion and alcohol.

\section{Conclusions}

This paper started with a review of the history of comparable corpora in 
\autoref{sec:history},
followed by a discussion of challenges for the future in \autoref{sec:future}.

\begin{enumerate}
    \setlength{\itemsep}{0pt}
    \setlength{\parskip}{0pt}
    \setlength{\parsep}{0pt}
    \item BLI is based on a (too) simple view of the lexicon.  Can we capture etymology?  Differences between monolingual and bilingual senses?
    \item 
     Transfer learning to growth languages: Avoid pivoting via English.  Better to prompt with pictures.  If we have to translate, it is better to translate into English than out of English to avoid imposing
    American values on others.
      \item Similarities between CC and recommender systems for academic search: can we compare and contrast a query document with candidate recommendations?  Can we cluster documents in monolingual and multilingual settings, and compare/contrast within and across clusters?
    \item Filter bubbles:
    chat bots currently lack a historian's ability to approach conflicts from multiple perspectives;
    bots made in America are trained on corpora from an American perspective with American biases.
    Can we capture ``possible worlds'' and diverse perspectives?
\end{enumerate}

\bibliography{custom}

\end{document}